\documentclass{article} 
\usepackage{colm2024_conference}

\usepackage{booktabs}
\usepackage{graphicx}
\usepackage{amsmath}
\usepackage{fixmath}
\usepackage{enumitem}
\usepackage{wrapfig}
\usepackage{algorithm}
\usepackage{algpseudocode}
\usepackage{natbib}
\usepackage{makecell}
\usepackage{booktabs}
\usepackage{array}
\usepackage{amssymb}
\usepackage{amsfonts}
\usepackage{multirow}
\usepackage{verbatim}
\usepackage{caption}
\usepackage{longtable}
\usepackage{supertabular}
\usepackage{CJKutf8}
\usepackage[utf8]{inputenc} 
\usepackage[T1]{fontenc}
\usepackage[french,vietnamese,mongolian,greek,english]{babel}
\usepackage{pifont}

\usepackage{enumitem}
\usepackage{tablefootnote}
\usepackage{xspace}
\usepackage{textcomp}
\usepackage{makecell}
\usepackage{lscape} 
\usepackage{siunitx}
\usepackage{listings}
\usepackage{xcolor}
\usepackage{svg}
\definecolor{dt}{gray}{0.7}
\definecolor{basecolor}{HTML}{4472C4}
\definecolor{rlcolor}{HTML}{ED7D31}
\usepackage{pifont}       
\usepackage{bbding}       
\usepackage{fontawesome}

\usepackage{scrextend}

\usepackage{tgpagella}
\usepackage{latexsym}
\usepackage[T1]{fontenc}
\usepackage[utf8]{inputenc}
\usepackage{microtype}
\definecolor{mydarkblue}{rgb}{0,0.08,0.45}
\definecolor{citecolor}{HTML}{0071BC}
\usepackage{url}            
\usepackage{nicefrac}       
\usepackage{changepage}
\usepackage{xargs}          
\usepackage{wrapfig,lipsum,booktabs}
\usepackage{longtable}
\usepackage{subcaption}
\usepackage{endnotes}

\usepackage{pgfplots}
\usetikzlibrary{pgfplots.groupplots}
\pgfplotsset{compat=1.3}
\usepackage{tikz}
\usetikzlibrary{patterns}

\usepackage{bm}

\renewcommand{\mathbf}[1]{\bm{#1}}

\usepackage[most]{tcolorbox}

\usepackage{hyperref}
\definecolor{darkblue}{rgb}{0, 0, 0.5}
\hypersetup{colorlinks=true, citecolor=darkblue, linkcolor=darkblue, urlcolor=darkblue, bookmarks=true, bookmarksopen=false, bookmarksnumbered=true}

\usepackage[capitalize,noabbrev]{cleveref}

\crefname{section}{Sec.}{Secs.}
\Crefname{section}{Sec.}{Secs.}
\crefname{subsection}{Sec.}{Secs.}
\Crefname{subsection}{Sec.}{Secs.}
\crefformat{section}{Sec.~#2#1#3}
\crefformat{subsection}{Sec.~#2#1#3}
\crefname{table}{Tab.}{Tabs.}
\Crefname{table}{Tab.}{Tabs.}
\crefname{figure}{Fig.}{Figs.}
\Crefname{figure}{Fig.}{Figs.}
\crefname{algorithm}{Algorithm}{}
\crefname{equation}{Eqn.}{Eqns.}
\Crefname{equation}{Eqn.}{Eqns.}
\crefformat{equation}{Eqn.~(#2#1#3)}
\crefname{appendix}{Appendix}{}
\usepackage{multicol}
\usepackage{tcolorbox}
\usepackage{titlesec}
\usepackage{float}
\titleformat*{\section}{\large\bfseries}
\usepackage{array} 
\newcolumntype{P}[1]{>{\centering\arraybackslash}p{#1}} 
\usepackage{multirow}

\usepackage{adjustbox}
\definecolor{objblue}{RGB}{3,139,221}  
\definecolor{attrred}{RGB}{255,67,67}    
\definecolor{easygreen}{RGB}{0,156,75}  
\definecolor{middleyellow}{RGB}{242,89,34}  
\definecolor{hardred}{RGB}{216,56,58}
\usepackage{colortbl}
\usepackage{array}

\usepackage[most]{tcolorbox}
\definecolor{BoxBackground}{RGB}{240, 240, 240} 
\definecolor{BoxFrame}{RGB}{0, 0, 0} 
\definecolor{TitleBackground}{RGB}{0, 0, 0} 
\definecolor{TitleText}{RGB}{255, 255, 255} 

\tcbset{
  academicbox/.style={
    boxsep=5pt,
    left=2pt,
    right=2pt,
    bottom=0.5pt,
    boxrule=0.5pt,
    colback=BoxBackground,
    colframe=BoxFrame,
    colbacktitle=TitleBackground,
    coltitle=TitleText,
    enhanced,
    attach boxed title to top left={yshift=-0.1in,xshift=0.1in},
    boxed title style={boxrule=0pt,colframe=white},
    title={#1},
  }
}
\newtcolorbox{AcademicBox}[1][]{academicbox=#1}

\title{Qwen-Image-2.0-RL Technical Report}

\author{  Yixian Xu\thanks{Equal contribution.} , Kaiyuan Gao$^*$, Yuxiang Chen$^*$, Yilei Chen,  Zecheng Tang, Zihao Liu, \\ Zikai Zhou, Deqing Li, Hao Meng, Kuan Cao, Jiahao Li, Jie Zhang, Liang Peng, \\ Lihan Jiang, Ningyuan Tang, Shengming Yin, Tianhe Wu, Xiaoyue Chen, Yan Shu, Yanran Zhang, Yi Wang, Yu Wu, Yujia Wu, Zekai Zhang, Zhendong Wang, \\Xiao Xu, Kun Yan, Chenfei Wu\thanks{Corresponding author.}}

\begin{document}

\maketitle
\begin{abstract}
We present \textbf{Qwen-Image-2.0-RL}, a post-training pipeline that applies reinforcement learning from human feedback (RLHF) and on-policy distillation (OPD) to improve both the visual quality and instruction-following capability of the Qwen-Image-2.0 diffusion model. To provide reliable reward signals, we construct task-specific composite reward models by fine-tuning vision-language models with a pointwise scoring paradigm and chain-of-thought reasoning. For text-to-image generation, the reward models cover alignment, aesthetics, and portrait fidelity dimensions. For image editing tasks, the reward system addresses instruction-following accuracy and face identity preservation. Building on this reward system, we develop a scalable GRPO-based RL training framework, incorporating a hybrid classifier-free guidance (CFG) strategy to preserve pre-trained knowledge, prompt curation via intra-group reward range filtering, and per-category reward weight calibration. To merge the task-specialized RL policies for T2I and editing, we propose on-policy distillation as the final training stage, which consolidates multiple teachers into a single student model through trajectory-level velocity matching. Extensive evaluation shows that Qwen-Image-2.0-RL achieves 57.84 overall score on Qwen-Image-Bench (+2.61 over the base model), Elo ratings of 1193 in text-to-image arena (+78) and 1349 in image edit arena (+93),  demonstrating consistent gains in aesthetic quality, prompt adherence, and editing accuracy.
\end{abstract}

\section{Introduction}
Diffusion and flow-based generative models~\citep{sohl2015deep,ho2020denoising,song2021score} have achieved remarkable success in high-fidelity image generation. The field has progressed to latent diffusion models~\citep{rombach2021highresolution,podell2023sdxl}  with scalable Transformer-based architectures~\citep{peebles2023scalable,chen2024pixartalpha,esser2024scaling,ma2024sit}. More recent systems~\citep{flux2024,flux-2-2025,wu2025qwen,cai2025z,Cao2025HunyuanImage3T} have further adopted vision-language foundation models as conditional encoders, whose stronger semantic grounding and multimodal world knowledge enable more precise instruction following and text-image alignment. Meanwhile, commercial systems~\citep{gao2025seedream,seedream2025seedream,gpt-image-1.5,nano-pro} have pushed the frontier of generation quality, and unified architectures~\citep{wu2025qwen,labs2025kontext} have extended these capabilities to image editing, enabling a single model to serve both generation and editing tasks within a shared framework.

Despite impressive pre-training results, a persistent gap remains between the outputs of supervised-trained diffusion models and human aesthetic expectations. Supervised training optimizes the denoising score matching objective that does not directly capture the human preference such as compositional harmony, texture, prompt faithfulness, and stylistic coherence. Reinforcement learning from human feedback (RLHF), which has achieved remarkable success in aligning large language models~\citep{shao2024deepseekmath}, offers a principled approach to close this gap. Based on reward signals that encode human preferences, the RLHF paradigm directly optimizes the model with respect to the reward signals.

However, extending the RLHF paradigm to diffusion models introduces distinct challenges. First, reliable reward signals must capture diverse quality dimensions across fundamentally different tasks such as text-to-image (T2I) generation and image editing, spanning global aesthetics and prompt adherence for T2I, and fine-grained identity preservation for editing. This necessitates a composite, task-aware reward design. Second, existing RL frameworks for diffusion models have primarily been validated under LoRA fine-tuning settings~\citep{liu2025flowgrpo,zheng2025diffusionnft,wang2025grpo}. Real-world scenarios involving multiple reward signals, diverse task types, and full-parameter training at scale remain underexplored. Third, practical deployment demands consolidating task-specialized RL policies into a single model without sacrificing per-task quality.

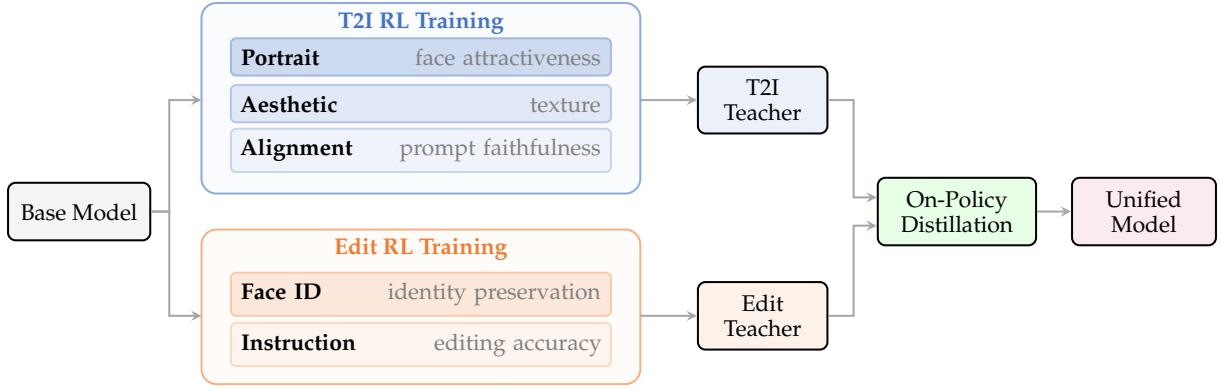
\begin{figure}[t]
\centering
\resizebox{\textwidth}{!}{%
\begin{tikzpicture}[>=stealth,
    box/.style={draw, rounded corners=3pt, thick, minimum height=0.8cm, align=center, font=\small, inner sep=5pt},
    rbar/.style={draw, rounded corners=2pt, minimum height=0.5cm, text width=5cm, font=\footnotesize, inner sep=4pt, thick},
    arr/.style={->, thick, color=gray!70},
]
  \node[box, fill=gray!8, minimum width=1.8cm] (base) at (0, 0) {Base Model};

  \draw[basecolor!60, thick, rounded corners=6pt, fill=basecolor!3]
    (1.7, 0.25) rectangle (7.8, 2.9);
  \node[basecolor, font=\small\bfseries] at (4.75, 2.62) {T2I RL Training};
  \node[rbar, fill=basecolor!8, draw=basecolor!30] (r_align) at (4.75, 0.85) {
    \textbf{Alignment} \hfill \textcolor{gray}{prompt faithfulness}};
  \node[rbar, fill=basecolor!16, draw=basecolor!40] (r_aes) at (4.75, 1.5) {
    \textbf{Aesthetic} \hfill \textcolor{gray}{texture}};
  \node[rbar, fill=basecolor!26, draw=basecolor!50] (r_port) at (4.75, 2.15) {
    \textbf{Portrait} \hfill \textcolor{gray}{face attractiveness}};

  \draw[rlcolor!60, thick, rounded corners=6pt, fill=rlcolor!3]
    (1.7, -2.4) rectangle (7.8, -0.25);
  \node[rlcolor, font=\small\bfseries] at (4.75, -0.52) {Edit RL Training};
  \node[rbar, fill=rlcolor!8, draw=rlcolor!30] (r_inst) at (4.75, -1.85) {
    \textbf{Instruction} \hfill \textcolor{gray}{editing accuracy}};
  \node[rbar, fill=rlcolor!18, draw=rlcolor!40] (r_face) at (4.75, -1.15) {
    \textbf{Face ID} \hfill \textcolor{gray}{identity preservation}};

  \node[box, fill=basecolor!10, minimum width=1.8cm] (t2it) at (9.5, 1.55) {T2I\\Teacher};
  \node[box, fill=rlcolor!10, minimum width=1.8cm] (editt) at (9.5, -1.45) {Edit\\Teacher};

  \node[box, fill=green!10, minimum width=2.2cm] (opd) at (12.2, 0) {On-Policy\\Distillation};

  \node[box, fill=purple!8, minimum width=2cm] (student) at (14.8, 0) {Unified\\Model};

  \draw[arr] (base.east) -- ++(0.25, 0) |- (1.7, 1.55);
  \draw[arr] (base.east) -- ++(0.25, 0) |- (1.7, -1.45);
  \draw[arr] (7.8, 1.55) -- (t2it.west);
  \draw[arr] (7.8, -1.45) -- (editt.west);
  \draw[arr] (t2it.east) -- ++(0.35, 0) |- (opd.170);
  \draw[arr] (editt.east) -- ++(0.35, 0) |- (opd.190);
  \draw[arr] (opd.east) -- (student.west);
\end{tikzpicture}%
}
\caption{Overview of the Qwen-Image-2.0-RL training pipeline. Starting from a shared base model, we train two task-specialized RL policies with dedicated reward compositions: T2I generation uses a layered reward design progressing from prompt faithfulness to texture quality to portrait-specific optimization, while editing focuses on instruction accuracy and identity preservation. The resulting teachers are merged into a unified model via on-policy distillation.}
\label{fig:pipeline}
\end{figure}

In this work, we address these challenges through a unified post-training pipeline (\cref{fig:pipeline}) applied to the Qwen-Image-2.0~\citep{zhao2026qwen} foundation model. Our contributions are summarized as follows:

\begin{enumerate}[leftmargin=*,itemsep=2pt]
    \item \textbf{VLM-based composite reward models (\cref{sec:reward}).} We construct task-specific reward suites by fine-tuning Qwen series VLM with chain-of-thought reasoning enabled, adopting a pointwise scoring paradigm that we find empirically superior to pairwise training. For T2I tasks, the rewards follow a layered design: an alignment reward ensures prompt faithfulness, an aesthetic reward improves texture and composition, and a portrait reward targets facial attractiveness. For image editing tasks, we combine an instruction-following reward with a dedicated face identity consistency scorer to capture global structural preservation and subtle facial identity shifts.

    \item \textbf{Scalable RL training framework (\cref{sec:training}).} We adopt a GRPO-based RL framework with multi-reward advantage computation~\citep{shao2024deepseekmath,liu2026gdpo}. In addition, we introduce a hybrid CFG strategy that applies guidance during rollout sampling but excludes it from the policy optimization objective, balancing training stability with preservation of pre-trained knowledge. We further propose a strategy to filter training prompts and per-category reward weight adjustment to balance optimization across various visual domains.

    \item \textbf{On-policy distillation (\cref{sec:opd}).} Once the RL policies are trained for specific tasks including T2I and image editing, we propose on-policy distillation (OPD) to unify task-specialized RL teachers into a single student model via trajectory-level velocity matching. OPD avoids cross-task optimization conflicts and eliminates reward model dependency.
\end{enumerate}

The resulting model, \textbf{Qwen-Image-2.0-RL}, achieves strong performance across multiple evaluation settings (\cref{sec:evaluation}): a 57.84 overall score on Qwen-Image-Bench (+2.61 over the base model), and Elo ratings of 1193 in text-to-image arena (+78) and 1349 in image edit arena (+93), demonstrating consistent improvements in aesthetic quality, prompt adherence, and editing accuracy for both T2I and image editing tasks.

\section{Backgrounds}

\subsection{Diffusion Models}
Diffusion models have become the dominant paradigm for high-fidelity image generation~\citep{sohl2015deep,ho2020denoising,song2021score,liu2022flow,lipman2022flow}. To model the high-dimensional data distribution $p_{\text{data}}$, diffusion models define a forward noising process that progressively corrupts data samples and learn a reverse process for generation. Following the Flow Matching framework~\citep{lipman2022flow,liu2022flow}, the forward path is defined by
\begin{align}
    \mathbf{x}_t=(1-t)\mathbf{x}_0+t\mathbf{\epsilon}, \quad t\in [0,1],
\end{align}
where $\mathbf{x}_0\sim p_{\text{data}}$ is a clean data sample and $\mathbf{\epsilon}\sim\mathcal{N}(\mathbf{0}, \mathbf{I})$ is standard Gaussian noise. The time derivative of this interpolation defines the conditional velocity field $ \mathbf{v} := \mathbf{\epsilon} - \mathbf{x}_0$.
A neural network $\mathbf{v}_\theta(\mathbf{x}_t, t,c)$ is trained to approximate this velocity field by minimizing the flow matching objective:
\begin{align}
    \mathcal{L}_{\text{FM}}(\theta) = \mathbb{E}_{t\sim p_t,\,\mathbf{x}_0\sim p_{\text{data}},\,\mathbf{\epsilon}\sim\mathcal{N}(\mathbf{0},\mathbf{I})} \left\| \mathbf{v}_\theta(\mathbf{x}_t, t,c) - (\mathbf{\epsilon} - \mathbf{x}_0) \right\|^2,
    \label{eq:fm_loss}
\end{align}
where $p_t$ is the distribution of the training timestep.
Once trained, samples are generated by solving the probability flow ordinary differential equation (ODE) backward from $t=1$ to $t=0$:
\begin{align}
    \frac{\mathrm{d}\mathbf{x}_t}{\mathrm{d}t} = \mathbf{v}_\theta(\mathbf{x}_t, t,c), \quad \mathbf{x}_1 \sim \mathcal{N}(\mathbf{0}, \mathbf{I}).
    \label{eq:sampling-ode}
\end{align}
\subsection{Reinforcement Learning for Diffusion Models}
\label{sec:rl_background}

Recent work has explored extending reinforcement learning to flow matching models for aligning generation quality with human preferences. Flow-GRPO~\citep{liu2025flowgrpo} extends Group Relative Policy Optimization (GRPO,~\citealt{shao2024deepseekmath}) to flow matching models by formulating the multi-step denoising trajectory as a Markov decision process (MDP). Given a prompt $c$, the policy $\pi_\theta$ generates a group of $G$ images $\{\mathbf{x}_0^{(1)}, \ldots, \mathbf{x}_0^{(G)}\}$, and a reward model $R(\mathbf{x}_0, c)$ evaluates each sample. The advantage of the $i$-th sample is computed by group-level normalization:
\begin{align}
    A(\mathbf{x}_0^{(i)}, c) = \frac{R(\mathbf{x}_0^{(i)}, c) - \mu_c}{\sigma_c},
    \label{eq:rl_adv}
\end{align}
where $\mu_c$ and $\sigma_c$ are the mean and standard deviation of rewards within the group. A central challenge is that flow matching relies on a deterministic ODE for generation, which hinders the direct application of GRPO. Flow-GRPO addresses this by using an equivalent stochastic sampler:
\begin{align}
    \mathrm{d}\mathbf{x}_t = \left[\mathbf{v}_\theta(\mathbf{x}_t, t) + \frac{\sigma_t^2}{2t}\left(\mathbf{x}_t + (1-t)\mathbf{v}_\theta(\mathbf{x}_t, t)\right)\right]\mathrm{d}t + \sigma_t \mathrm{d}\mathbf{w}_t,
    \label{eq:flow_sde}
\end{align}
where $\sigma_t\geq0$ controls the noise scale and $\mathbf{w}_t$ denotes the standard Wiener process. Under Euler-Maruyama discretization, the transition density becomes Gaussian, enabling tractable computation of the importance sampling ratio $r_t^{(i)}(\theta) = \pi_\theta(\mathbf{x}_{t-1}^{(i)} | \mathbf{x}_t^{(i)}, c) / \pi_{\theta_\text{old}}(\mathbf{x}_{t-1}^{(i)} | \mathbf{x}_t^{(i)}, c)$.
The policy is then optimized via a clipped surrogate objective:
\begin{align}
    \mathcal{L}_{\text{Flow-GRPO}}(\theta) = -\mathbb{E}_{c\sim\mathcal{D},\,\mathbf{x}_{0:T} \sim \pi_{\theta_{\text{old}}}(\cdot | c)}\frac{1}{G} \sum_{i=1}^{G} \frac{1}{T} \sum_{t=0}^{T-1} \left[\min\!\left(r_t^{(i)}(\theta)\,A(\mathbf{x}_0^{(i)}, c) ,\; \hat{r}_t^{(i)}(\theta)A(\mathbf{x}_0^{(i)}, c) \right) \right],
    \label{eq:flow_grpo_obj}
\end{align}
where $\hat{r}_t^{(i)}(\theta):=\text{clip}\!\left(r_t^{(i)}(\theta), 1{-}\epsilon, 1{+}\epsilon\right)$ is the clipped version of importance sampling ratio. 

DiffusionNFT~\citep{zheng2025diffusionnft} proposes an alternative formulation that uses the forward diffusion process for policy optimization. For a noisy state $\mathbf{x}_t=(1-t)\mathbf{x}_0+t\mathbf{\epsilon}$, three velocity predictions are computed: the current policy $\mathbf{v}_\theta(\mathbf{x}_t, t,c)$, the old policy $\mathbf{v}_{\theta_{\text{old}}}(\mathbf{x}_t, t,c)$, and the reference policy $\mathbf{v}_{\theta_{\text{ref}}}(\mathbf{x}_t, t,c)$. The method constructs positive and negative velocity predictions defined by
\begin{align}
    \mathbf{v}^{+}_\theta = \beta \cdot \mathbf{v}_\theta + (1-\beta) \cdot \mathbf{v}_{\theta_{\text{old}}}, \quad
    \mathbf{v}^{-}_\theta = (1+\beta) \cdot \mathbf{v}_{\theta_{\text{old}}} - \beta \cdot \mathbf{v}_\theta,
\end{align}
where $\beta$ is the interpolation strength. Then the training objective of DiffusionNFT is given by
\begin{align}
    \mathcal{L}_{\text{NFT}} = \mathbb{E}_{c\sim\mathcal{D},\,\mathbf{x}_0 \sim \pi_{\theta_{\text{old}}}(\cdot | c),\, \mathbf{\epsilon}\sim\mathcal{N}(\mathbf{0},\mathbf{I})}\left[r(\mathbf{x}_0, c) \Vert \mathbf{v}^{+}_\theta(\mathbf{x}_t, t,c)-\mathbf{v}\Vert^2+(1-r(\mathbf{x}_0, c)) \Vert \mathbf{v}^{-}_\theta(\mathbf{x}_t, t,c)-\mathbf{v}\Vert^2\right],
    \label{eq:nft-loss}
\end{align}
where $\mathbf{x}_t=(1-t)\mathbf{x}_0+t\mathbf{\epsilon}$, $\mathbf{v}=\epsilon-\mathbf{x}_0$, $r=\frac{\text{clip}(A, -A_{\max}, A_{\max})}{2 A_{\max}} + 0.5\in[0,1]$ is a rescaled version of the group-relative advantage. 
To prevent the policy from deviating too far from the pre-trained reference, a KL penalty is added:
\begin{align}
    \mathcal{L}_{\text{KL}} = \mathbb{E}_{c\sim\mathcal{D},\,\mathbf{x}_0 \sim \pi_{\theta_{\text{old}}}(\cdot | c),\, \mathbf{\epsilon}\sim\mathcal{N}(\mathbf{0},\mathbf{I})}\left\| \mathbf{v}_\theta(\mathbf{x}_t, t,c) - \mathbf{v}_{\theta_{\text{ref}}}(\mathbf{x}_t, t,c) \right\|^2.
\end{align}

\section{Reward Modeling}
\label{sec:reward}
The reward signal capturing human preference is the primary component of the RL training. To align the pretrained image generative model with the human preferences, we construct task-specific composite reward models for different evaluation dimensions. Our reward system combines VLM-based scorers for semantic and aesthetic assessment with model-based scorers for fine-grained identity preservation, tailored to both T2I and image editing tasks.

\subsection{Reward Model Training Paradigms}
\label{sec:reward_paradigm}
\begin{figure}[t]
    \makebox[\linewidth]{
        \includegraphics[width=\linewidth]{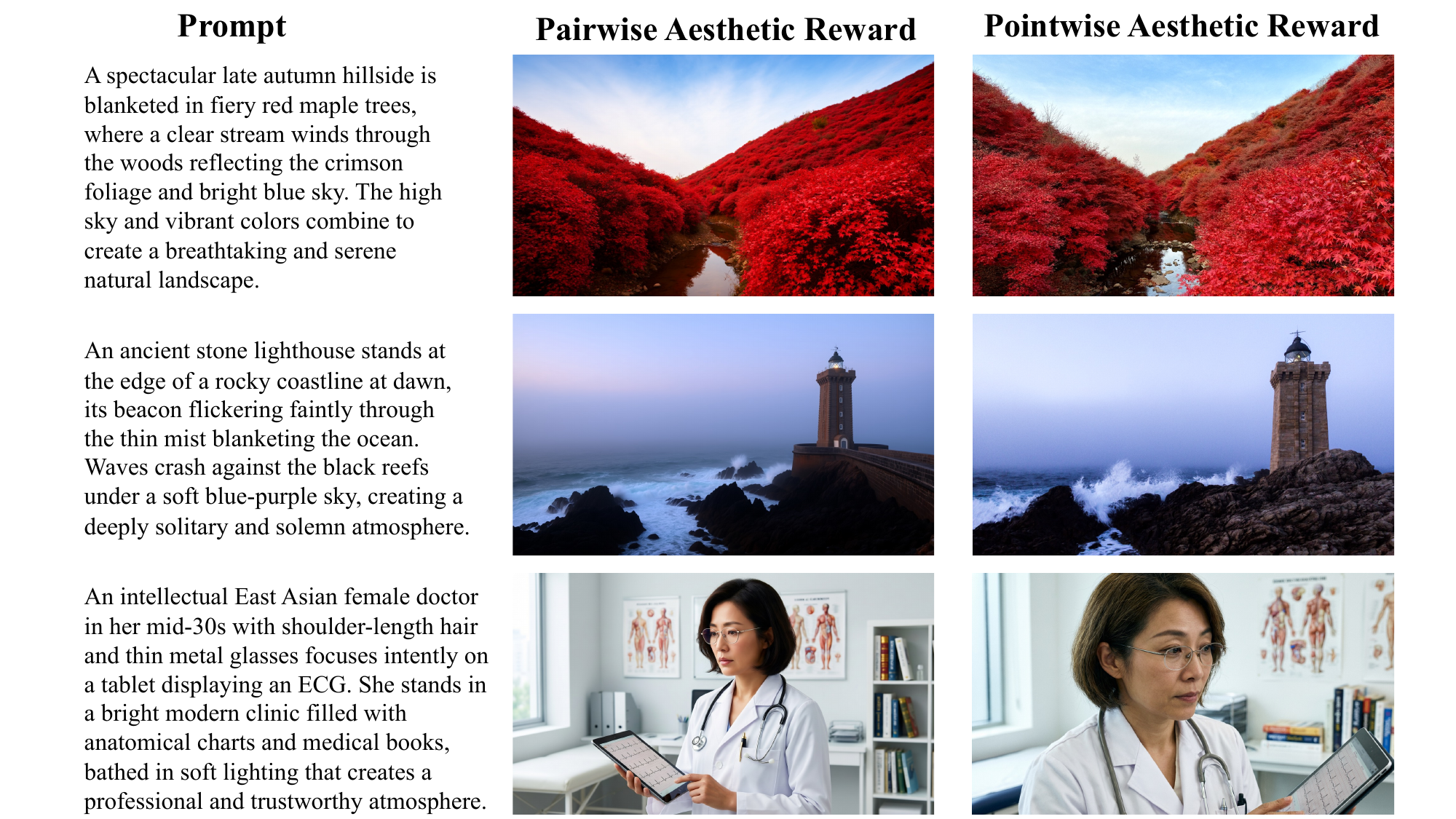}
    }
    \caption{Qualitative comparison of RL training outcomes using pointwise vs.\ pairwise reward model training paradigms. Both reward models target the same evaluation dimensions (aesthetic quality and visual texture) and are trained on data from the same model pool, isolating the effect of the training paradigm. The pointwise-trained reward model produces images with consistently better visual quality, finer texture detail, and fewer artifacts.}
    \label{fig:reward_comparison}
\end{figure}

We explore two training paradigms for reward model fine-tuning. The first is \textbf{pairwise reward training}, where the model is optimized on pairs of images $(\mathbf{x}_w, \mathbf{x}_l)$ generated from the same prompt $c$, with $\mathbf{x}_w$ preferred by human annotators over $\mathbf{x}_l$. The Vision-Language Model (VLM) produces scalar scores $R_\phi(x, c)$ for each image, and the training objective minimizes the Bradley-Terry ranking loss:
\begin{align}
    \mathcal{L}_{\text{pair}} = -\sum_{(\mathbf{x}_w, \mathbf{x}_l, c)} \log \sigma\!\left(R_\phi(\mathbf{x}_w, c) - R_\phi(\mathbf{x}_l, c)\right),
\end{align}
where $\sigma(x)=\frac{1}{1+e^{-x}}$ is the sigmoid function.
The second is \textbf{pointwise reward training}, where each image $\mathbf{x}$ is paired with an absolute human-annotated score $y \in \mathbb{R}$, and the model is trained to directly regress to this score:
\begin{align}
    \mathcal{L}_{\text{point}} = \sum_{(\mathbf{x}, c, y)} \left(R_\phi(\mathbf{x}, c) - y\right)^2.
\end{align}
In practice, the reward model is trained to output tokens in discrete score set $\mathcal{S}=\{1,2,3,4,5\}$, and the reward score is produced by the expectation under the VLM's probability distribution $p_\phi$:
\begin{align}
    R_\phi(\mathbf{x}, c)=\sum_{s\in \mathcal{S}}s\cdot p_\phi(s|\mathbf{x},c).
\end{align}
To compare the two training paradigms, we construct two annotated datasets that deliberately share the same evaluation focus---aesthetic quality and visual texture---while differing only in annotation format. Both datasets draw images from a common pool of state-of-the-art AIGC models, ensuring that the comparison isolates the effect of the training paradigm rather than data distribution.

\paragraph{Pointwise annotation.} We collect images datasets with prompts randomly sampled from high-quality portrait reference images and rewritten to diversify linguistic patterns beyond the training distribution. Each image is independently scored by human annotators on a 5-point Likert scale along two structured dimensions: (1) Quality---assessing clarity, lighting, color balance, stylistic coherence, and material texture; and (2) Fidelity---evaluating structural correctness, physical consistency, and absence of AI artifacts such as unnatural smoothing or texture repetition.

\paragraph{Pairwise annotation.} We collect image pairs dataset, where two images generated from the same prompt are presented side by side for preference judgment. Annotation follows a strict priority hierarchy: image-text consistency $>$ structural distortion $>$ texture quality $>$ aesthetic appeal. Under this scheme, when both images faithfully depict the prompt content and are free of structural distortions, which is the common case for images from high-quality models, the comparison reduces to a direct judgment of texture quality and aesthetic appeal. Each pair is further labeled with auxiliary attributes including sample validity, text distortion presence, and human figure distortion presence to support fine-grained analysis.

\paragraph{Comparison.} We train two reward models by finetuning the same VLM architectures and use each as the RL training reward. As shown in \cref{fig:reward_comparison}, the pointwise-trained reward model produces images with consistently better visual quality and fewer artifacts. We attribute this to the richer supervisory signal of absolute scores: pointwise annotations encode how good an image is on a calibrated scale, whereas pairwise annotations only capture which is better. Based on this finding, we adopt the pointwise paradigm as the default training objective for all VLM-based reward models in the final system.

\subsection{Reward Models for Text-to-Image Generation}
\label{sec:reward_t2i}

Having established the pointwise paradigm as our default training objective (\cref{sec:reward_paradigm}), we now describe the specific reward models for T2I generation. All VLM-based T2I reward models are implemented by fine-tuning Qwen series VLM. Our T2I reward model design follows a layered logic. The most fundamental requirement for image generation is faithfulness to the user's prompt: a visually stunning image that ignores the specified content is a failed generation. We therefore begin with an image-text alignment reward that evaluates semantic correspondence without considering aesthetics. Once prompt adherence is established, we layer on an aesthetic reward to enrich texture fidelity. Finally, because human-subject images demand more than generic aesthetic quality, we introduce a dedicated portrait reward that specifically optimizes facial attractiveness and fine-grained skin and hair realism.

\paragraph{Image-text alignment reward.} This is the most fundamental reward, measuring the semantic correspondence between the generated image and the input prompt. It explicitly penalizes outputs that omit, misinterpret, or contradict user-specified requirements, without considering aesthetic merit. The VLM evaluates prompt adherence along a priority hierarchy: (1) object presence and count accuracy, (2) attribute correctness (color, size, shape, material), (3) spatial relationship fidelity, and (4) action and pose accuracy. Images that fail the highest-priority criteria are capped at low scores regardless of other qualities.

\paragraph{Aesthetic reward.} Building upon prompt-faithful generation, this reward assesses the intrinsic visual quality of generated images, emphasizing compositional balance, realistic illumination, texture fidelity, and overall artistic coherence. The aesthetic reward model is trained on the pointwise-annotated dataset described in \cref{sec:reward_paradigm}.

\paragraph{Portrait reward.} Generic aesthetic optimization is insufficient for human-subject generation, where facial attractiveness and anatomical correctness are critical. This reward provides a specialized signal for improving facial proportion accuracy, identity-preserving facial details, and fine-grained skin and hair texture realism. The scorer explicitly checks for common failure modes such as incorrect finger counts, distorted facial features, and unnatural body proportions. We train this reward model on a separate portrait-specific dataset with prompts sampled from high-quality portrait reference images. The annotation rubric focuses specifically on facial attractiveness, skin and hair texture realism, capturing the standards of human beauty that generic aesthetic scoring cannot adequately address.

\subsection{Reward Models for Text-Guided Image Editing}
\label{sec:reward_edit}

Building on the success of the pointwise reward paradigm established for T2I, we transfer the same training methodology to image editing tasks. The VLM-based rewards are likewise built by fine-tuning Qwen series VLM, adapting the evaluation rubrics to editing-specific requirements. We additionally introduce a model-based face identity consistency scorer to address fine-grained identity preservation beyond the VLM's capabilities.

\paragraph{Instruction-following reward.} This reward evaluates whether user-specified modifications are accurately executed, covering editing operations such as object replacement, attribute modification, and style transfer. The fine-tuned VLM model receives the source image, the editing instruction, and the output image as a triplet, and is prompted to decompose the instruction into core editing requirements and non-core auxiliary requirements. The evaluation follows a structured rubric that assesses: (1) whether the core editing instruction has been fulfilled, (2) whether non-core requirements are addressed, and (3) whether the overall output is visually coherent.

\paragraph{Face identity consistency reward.} While the VLM-based visual consistency reward captures global structural preservation, we find that it is insufficient for reliably detecting subtle facial identity shifts. We therefore introduce a dedicated model-based face identity scorer. This model-based reward provides a precise, embedding-level identity preservation signal that complements the VLM's higher-level semantic consistency assessment.

\section{Training}
\label{sec:training}

With the reward models established in \cref{sec:reward} and the GRPO-based RL framework introduced in \cref{sec:rl_background}, we now describe our training pipeline. We first train separate RL policies for T2I generation and image editing using their respective reward compositions (\cref{sec:reward_t2i,sec:reward_edit}), then merge the resulting task-specialized models into a deployable model via on-policy distillation. In \cref{sec:pipeline}, we present the shared pipeline infrastructure for T2I and editing task. \cref{sec:task_opt} details the task-specific optimization choices. Finally, \cref{sec:opd} introduces on-policy distillation, which unifies the two task-specialized teachers into a single student model through trajectory-level velocity matching.

\subsection{Training Pipeline}
\label{sec:pipeline}

\begin{figure}[t]
    \makebox[\linewidth]{
        \includegraphics[width=\linewidth]{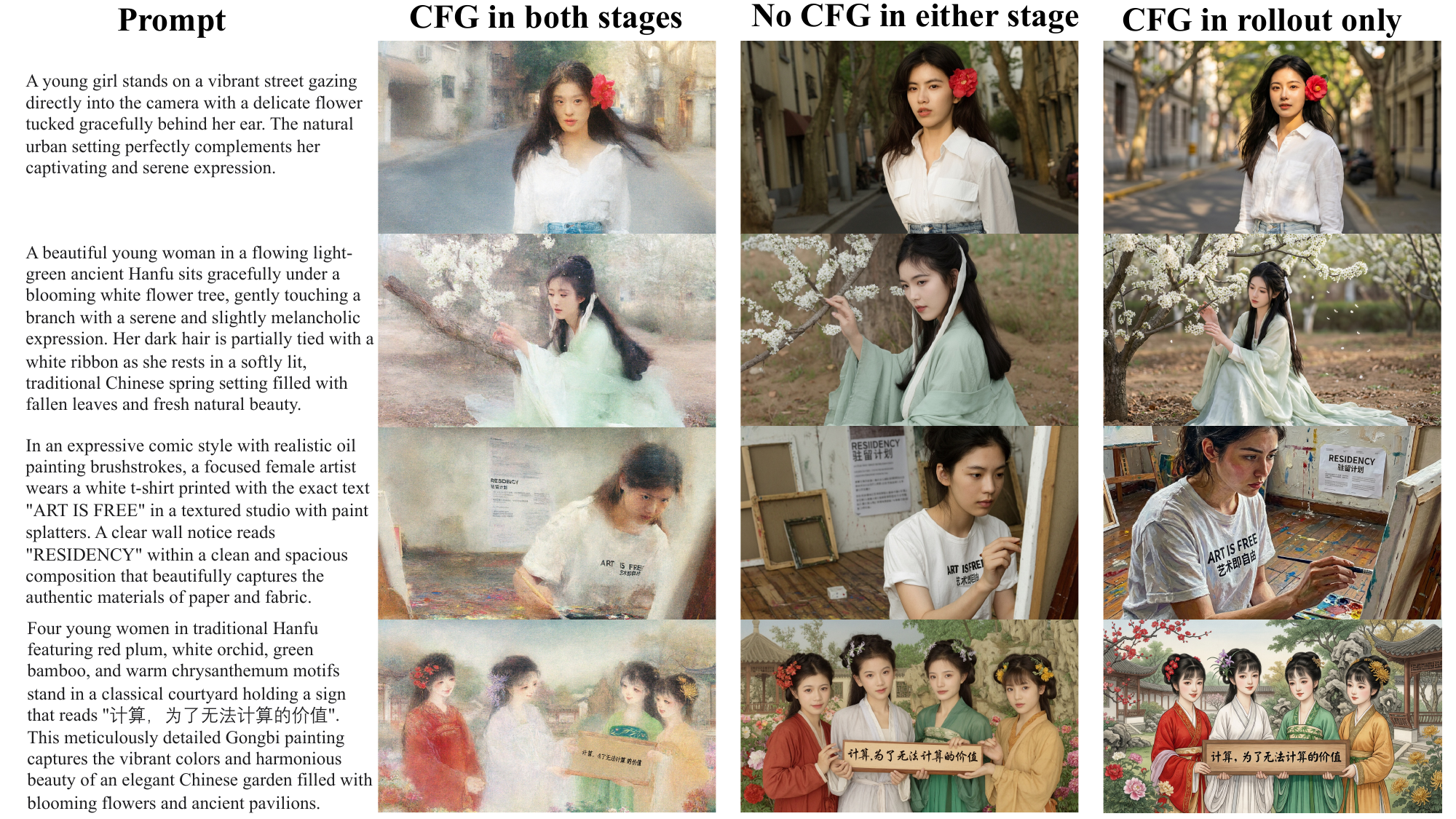}
    }
    \caption{Qualitative comparison of three CFG strategies during RL training. \textbf{Left}: CFG applied in both rollout and training leads to training instability and eventual image collapse. \textbf{Middle}: removing CFG from both stages causes progressive loss of stylization ability and world knowledge. \textbf{Right}: CFG in rollout only (our hybrid strategy) maintains training stability while preserving the pre-trained model's full generative capabilities.}
    \label{fig:cfg_comparison}
\end{figure}

\paragraph{Hybrid CFG strategy.} A key design consideration in diffusion-based RL is whether classifier-free guidance (CFG)~\citep{ho2022classifier} should be employed during rollout sampling and policy optimization. We systematically evaluate three strategies (see \cref{fig:cfg_comparison}):
\begin{itemize}
    \item CFG in both rollout and training. Applying CFG during both rollout sampling and the training stage leads to severe training instability. As training progresses, the generated images deteriorate completely, ultimately collapsing into incoherent outputs;
    \item No CFG in either stage. Although reward scores steadily improve, the model progressively loses stylization ability and world knowledge, failing to reproduce well-known celebrity appearances and losing the capacity for style-specific generation. We attribute this phenomenon to the base model's reliance on CFG to fully express its pre-trained knowledge during inference;
    \item CFG in rollout only. CFG is used during the rollout stage to generate high-quality candidates for reward evaluation, while the unconditional branch is excluded from the policy optimization objective.
\end{itemize}
We adopt the third, hybrid strategy. The CFG-guided rollout fully leverages the pre-trained model's capabilities to produce structurally coherent images that yield reliable reward signals. Meanwhile, the CFG-free training objective avoids the optimization difficulties introduced by jointly optimizing the conditional and unconditional branches, maintaining stable gradient updates and substantially reducing computational overhead.

\paragraph{Asynchronous reward pipeline.} Our reward models are deployed as remote API services, separate from the training process. Since reward scoring involves network I/O to these remote VLM endpoints, synchronous evaluation would bottleneck the training loop. We therefore decouple reward computation from model training through an asynchronous pipeline: after the policy generates a batch of images via GPU inference and gathers them across ranks, a background thread asynchronously submits the images to the remote reward API endpoints. Once the asynchronous reward responses return, all ranks synchronize to gather the raw scores, perform per-prompt-group normalization, and compute advantages for the policy gradient update. This design hides nearly all reward latency behind inference computation, enabling efficient scaling to multiple reward models without proportional increases in training time.

\paragraph{Multi-reward advantage computation.} As mentioned in \cref{sec:reward}, we use multiple reward models for the training of T2I and image editing task respectively. Inspired by \citet{liu2026gdpo}, the group-relative advantage in \cref{eq:rl_adv} is calculated by weighted summation with per-prompt-group normalization:
\begin{align}
    A(\mathbf{x}_0^{(i)}, c) = \sum_{k=1}^{K} w_k \cdot \frac{R_k(\mathbf{x}_0^{(i)}, c) - \mu_k}{\sigma_k},
    \label{eq:reward_composition}
\end{align}
where $R_k$ is the $k$-th reward model, $w_k$ is its weight satisfying $\sum_{k=1}^K w_k=1$, and $\mu_k$, $\sigma_k$ are the mean and standard deviation of $R_k$ computed within each prompt group. This per-prompt-group normalization is critical: it ensures that the composite reward is invariant to absolute scale differences across reward models, preventing any single reward dimension from dominating the advantage signal due to its numerical range. 

\subsection{Task-Specific Optimization}
\label{sec:task_opt}

\paragraph{Timestep sampling.} During rollout, images are generated using a 40-step ODE solver. A naive approach would apply the RL training objective at all 40 timesteps. However, we observe that this leads to rapid reward hacking, resulting in degradation within a few iterations. To address this issue, we train on only a subset of the rollout timesteps, with a particular emphasis on high-noise timesteps (i.e., those closer to $t=1$). High-noise timesteps govern global structure and semantic layout, making them more robust targets for policy optimization. By restricting the training signal to a carefully selected subset, we slow the reward exploitation process and ensure that the model improves comprehensively across quality dimensions.

\paragraph{Prompt curation.} Not all prompts contribute equally to policy improvement. We employ the trained reward models to filter the prompt pool before RL training. For each candidate prompt, the base model performs $G$ rollouts and the composite reward is computed for each sample. We then compute the intra-group range (maximum minus minimum reward) within each prompt group. Only prompts whose range exceeds a predefined threshold are retained for training. Prompts with uniformly high or low rewards across all samples provide a weak signal for policy optimization. This filtering step significantly improves training efficiency by concentrating compute on prompts where the policy has room for meaningful improvement.

\paragraph{Per-category reward calibration.} We organize the retained prompts into semantic categories (e.g., portrait, landscape, typography, general) and assign category-specific reward weight vectors. For instance, portrait prompts receive higher weight on the portrait fidelity reward, while typography prompts emphasize alignment accuracy. This per-category calibration ensures that the RL objective reflects the distinct quality requirements of each visual domain, preventing the optimization from converging to a single dominant style at the expense of others.

\subsection{On-Policy Distillation}
\label{sec:opd}

While the preceding RL optimization produces models with superior quality on each task, the resulting policies are task-specialized: a T2I-optimized model may exhibit degraded editing performance, and vice versa. To address this limitation, we propose On-Policy Distillation (OPD), which unifies multiple task-specialized RL-trained teachers into a single student model via trajectory-level velocity matching.

\paragraph{Training objective.} The student model $\mathbf{v}_\theta$ (initialized from the pre-trained base model) generates images by solving the reverse ODE from $t=1$ to $t=0$ with $N$ discrete steps, using a timestep schedule $\{t_0 = 1, t_1, \ldots, t_N \approx 0\}$. The full trajectory of the student model $\{\mathbf{x}_{t_0}, \mathbf{x}_{t_1}, \ldots, \mathbf{x}_{t_N}\}$ is saved, where $\mathbf{x}_{t_0} \sim \mathcal{N}(\mathbf{0}, \mathbf{I})$ is the initial noise sample and $\mathbf{x}_{t_N}$ is the denoised output. The student is trained to match the task-appropriate teacher's velocity at each point along its own trajectory:
\begin{align}
    \mathcal{L}_{\text{OPD}} = \mathbb{E}_{c,\mathbf{x}_{[1:N]}\sim \pi_\theta(\cdot|c)}\left[\sum_{n =1}^N \left\Vert \mathbf{v}_\theta(\mathbf{x}_{t_n}, t_n,c) - \mathbf{v}_{\theta^*}(\mathbf{x}_{t_n}, t_n,c) \right\Vert^2\right],
    \label{eq:opd_loss}
\end{align}
where $\mathbf{v}_{\theta^\star}$ is the task-related teacher model (see \cref{app:opd_derivation} for a formal derivation). The OPD objective ensures that the student learns to correct its own prediction errors on its own inference trajectories.

\paragraph{Multi-Teacher Distillation.} A central advantage of OPD is its ability to distill from multiple task-specialized teachers into a single student. We maintain two teacher models: a T2I teacher optimized for text-to-image generation with aesthetic, alignment, and portrait rewards, and an editing teacher optimized for image editing tasks with instruction-following and face identity preservation rewards. For each training batch, the appropriate teacher is selected based on the task type of the current sample. To manage GPU memory, only the active teacher is loaded onto the GPU at any time; inactive teachers are offloaded to CPU. This dynamic teacher activation mechanism allows training with multiple large teacher models without proportionally increasing GPU memory requirements. As teacher models are originally trained with CFG, we apply CFG during the teacher's velocity prediction in OPD, but keep the student model without CFG. By distilling from specialized teachers rather than jointly training with competing rewards, OPD avoids the optimization conflicts that arise when T2I and editing objectives are optimized simultaneously. In addition, the CFG is also integrated into the student model after OPD.

\begin{figure}[t]
    \makebox[\linewidth]{
        \includegraphics[width=\linewidth]{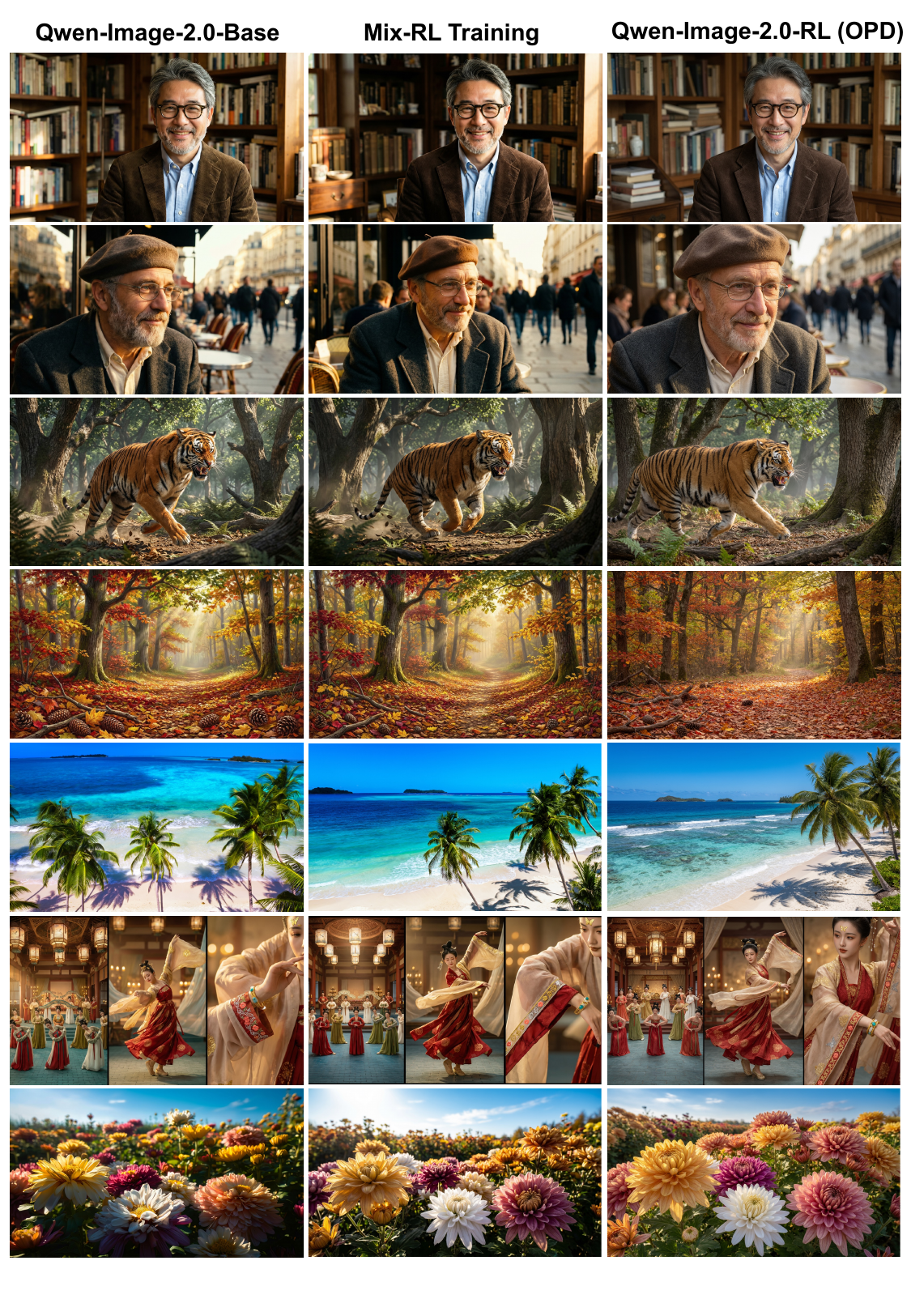}
    }
    \caption{Qualitative comparison across T2I generation scenarios among three model variants: pre-trained Qwen-Image-2.0-Base, Mix-RL (jointly trained on T2I and editing tasks with mixed RL rewards), and Qwen-Image-2.0-RL (task-specialized RL teachers distilled via on-policy distillation). The progression Qwen-Image-2.0-Base $\rightarrow$ Mix-RL $\rightarrow$ Qwen-Image-2.0-RL demonstrates that RL training improves visual quality over the pre-trained baseline, and that OPD further surpasses mixed RL by avoiding cross-task optimization conflicts.}
    \label{fig:qualitative}
\end{figure}

\begin{figure}[t]
    \makebox[\linewidth]{
        \includegraphics[width=\linewidth]{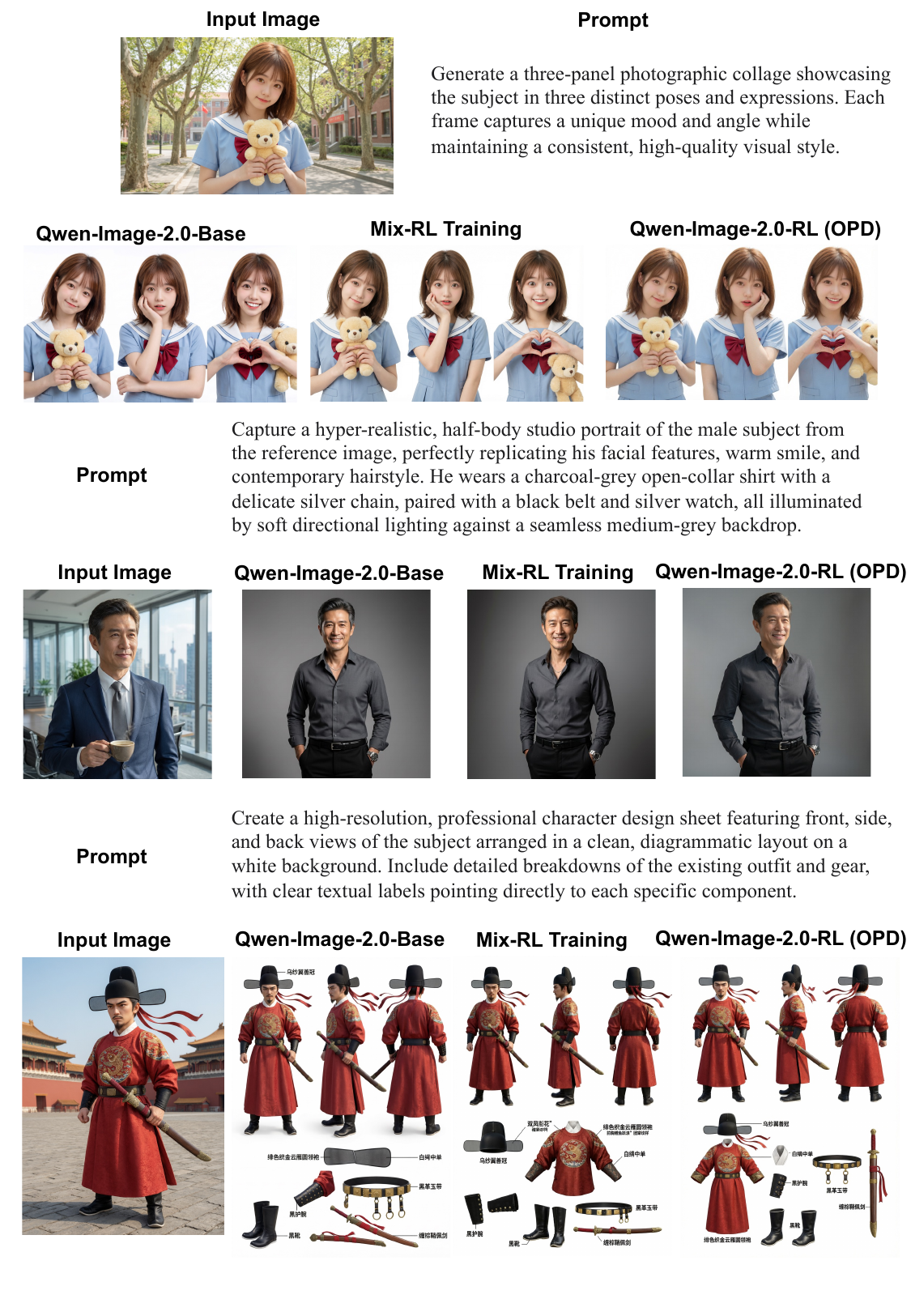}
    }
    \caption{Qualitative comparison across portrait editing scenarios among three model variants. Qwen-Image-2.0-RL achieves the best face identity preservation and instruction-following accuracy, while Mix-RL improves over the base model but still exhibits noticeable identity drift or incomplete edits under complex instructions.}
    \label{fig:qualitative_edit}
\end{figure}

\paragraph{Comparison with mixed RL training.} A natural alternative to the decomposed OPD pipeline is to train a single model with RL on mixed T2I and editing data (Mix-RL), where all task-specific rewards are jointly optimized in one training process. While this approach is simpler, it forces the model to simultaneously satisfy competing optimization objectives from different tasks, leading to suboptimal trade-offs. \cref{fig:qualitative} presents a three-way qualitative comparison across T2I scenarios among the pre-trained base model (Qwen-Image-2.0-Base), the Mix-RL baseline, and our final Qwen-Image-2.0-RL model produced via OPD. The comparison reveals a clear quality progression: Mix-RL already improves over the base model in texture fidelity, compositional coherence, and overall realism, confirming that RL training with our reward suite effectively enhances generation quality. However, Qwen-Image-2.0-RL consistently outperforms Mix-RL, producing sharper details, more accurate prompt adherence, and better aesthetic quality. A similar trend is observed for image editing: \cref{fig:qualitative_edit} shows that Qwen-Image-2.0-RL achieves superior face identity preservation and instruction-following accuracy compared to both the base model and Mix-RL, where the latter still suffers from identity drift or incomplete edits under complex instructions. This demonstrates the advantage of our decomposed strategy over jointly optimizing all task rewards in a single RL training process.

\section{Evaluation}
\label{sec:evaluation}

We evaluate Qwen-Image-2.0-RL from automated quality metrics on standardized benchmarks to human preference rankings on competitive arenas.

\paragraph{Text-to-image generation results.} We assess the effectiveness of our RL training framework on T2I using Qwen-Image-Bench~\citep{li2026qwen}, a creator-centric benchmark designed to evaluate T2I models across five first-level pillars: Quality, Aesthetics, Alignment, Real-world Fidelity, and Creative Generation. Evaluation is conducted by Q-Judger, a unified judge model trained on over 130K human-labeled image-prompt pairs annotated by 80 professional artists. 
\cref{tab:qwen_image_bench} presents the performance of Qwen-Image-2.0 before and after RL training alongside strong baselines on Qwen-Image-Bench. RL training yields consistent improvements across all five evaluation pillars, raising the overall score from 55.23 to 57.84. The most substantial gains emerge in Creative Generation (6.72 improvement) and Real-world Fidelity (4.29 improvement).

\begin{table}[t]
\centering
\caption{Performance comparison on Qwen-Image-Bench~\citep{li2026qwen}. Scores are on a $[0,100]$ scale, aggregated bottom-up from 56 third-level facets through a three-level taxonomy. Baseline models are sorted by overall score in ascending order. }
\label{tab:qwen_image_bench}
\resizebox{\textwidth}{!}{
\begin{tabular}{l c c c c c c}
\toprule
\textbf{Model} & \textbf{Quality} & \textbf{Aesthetics} & \textbf{Alignment} & \textbf{Real-world Fidelity} & \textbf{Creative Gen.} & \textbf{Overall} \\
\midrule
GLM Image & 49.26 & 50.64 & 47.90 & 44.69 & 45.23 & 48.19 \\
Kling Image 2.1 & 49.11 & 50.15 & 49.18 & 44.74 & 44.67 & 48.26 \\
Qwen Image & 48.44 & 52.25 & 50.72 & 43.16 & 47.30 & 49.23 \\
Imagen 4.0 & 50.16 & 52.68 & 51.64 & 44.84 & 47.94 & 50.29 \\
HunyuanImage 3.0 & 50.35 & 53.57 & 52.00 & 44.31 & 49.12 & 50.81 \\
Imagen 4.0 Ultra & 50.90 & 54.25 & 54.02 & 45.59 & 51.14 & 51.99 \\
Qwen Image 2512 & 51.76 & 54.74 & 52.72 & 47.00 & 50.19 & 52.06 \\
GPT Image 1 & 52.34 & 55.09 & 56.28 & 48.14 & 55.78 & 54.07 \\
FLUX 2 Pro & 52.30 & 56.94 & 57.01 & 47.29 & 56.18 & 54.57 \\
FLUX 2 Max & 53.64 & 56.85 & 57.35 & 49.35 & 56.50 & 55.33 \\
Seedream 4.0 & 54.01 & 58.81 & 56.64 & 51.05 & 58.15 & 56.21 \\
Seedream 4.5 & 54.41 & 58.72 & 57.31 & 51.69 & 60.64 & 56.78 \\
Seedream 5.0 & 52.55 & 58.40 & 58.90 & 51.92 & 65.29 & 57.22 \\
Nano Banana Pro & 55.67 & 60.26 & 61.25 & 54.07 & 66.23 & 59.45 \\
GPT Image 1.5 & 55.14 & 60.88 & 61.72 & 53.95 & 66.35 & 59.65 \\
Nano Banana 2.0 & 54.77 & 61.08 & 62.40 & 54.28 & 67.05 & 59.82 \\
GPT Image 2 & \textbf{58.65} & \textbf{67.53} & \textbf{65.85} & \textbf{57.38} & \textbf{75.23} & \textbf{64.69} \\
\midrule
Qwen-Image-2.0-Base & 52.29 & 57.10 & 57.64 & 47.54 & 58.22 & 55.23 \\
\textbf{Qwen-Image-2.0-RL} & \textbf{54.39} & \textbf{58.67} & \textbf{59.28} & \textbf{51.83} & \textbf{64.94} & \textbf{57.84} \\
\bottomrule
\end{tabular}
}
\end{table}

\begin{figure}[t]
\centering
\begin{subfigure}[b]{0.46\textwidth}
\centering
\begin{tikzpicture}[scale=0.58]
  \foreach \rr/\elolabel in {0.75/1050, 1.50/1150, 2.25/1250, 3.00/1350} {
    \draw[gray!25, thin]
      (90:\rr) -- (45:\rr) -- (0:\rr) -- (315:\rr) --
      (270:\rr) -- (225:\rr) -- (180:\rr) -- (135:\rr) -- cycle;
    \node[gray!50, font=\tiny, anchor=south west] at (67.5:\rr) {\elolabel};
  }
  \foreach \angle in {90,45,0,315,270,225,180,135} {
    \draw[gray!40, thin] (0,0) -- (\angle:3.2);
  }
  \node[font=\scriptsize, anchor=south] at (90:3.55) {Product};
  \node[font=\scriptsize, anchor=south west] at (45:3.4) {3D};
  \node[font=\scriptsize, anchor=west] at (0:3.4) {Cartoon};
  \node[font=\scriptsize, anchor=north west] at (315:3.45) {Photorealism};
  \node[font=\scriptsize, anchor=north] at (270:3.55) {Art};
  \node[font=\scriptsize, anchor=north east] at (225:3.4) {Portraits};
  \node[font=\scriptsize, anchor=east] at (180:3.55) {Text Rendering};
  \node[font=\scriptsize, anchor=south east] at (135:3.4) {Edit};
  \draw[basecolor, thick, fill=basecolor, fill opacity=0.15]
    (90:1.245) -- (45:1.073) -- (0:1.178) -- (315:1.343) --
    (270:1.193) -- (225:1.410) -- (180:1.253) -- (135:2.295) -- cycle;
  \foreach \angle/\rr in {90/1.245, 45/1.073, 0/1.178, 315/1.343, 270/1.193, 225/1.410, 180/1.253, 135/2.295} {
    \fill[basecolor] (\angle:\rr) circle (2pt);
  }
  \draw[rlcolor, thick, fill=rlcolor, fill opacity=0.15]
    (90:1.740) -- (45:1.770) -- (0:1.770) -- (315:2.025) --
    (270:1.665) -- (225:1.958) -- (180:1.838) -- (135:2.993) -- cycle;
  \foreach \angle/\rr in {90/1.740, 45/1.770, 0/1.770, 315/2.025, 270/1.665, 225/1.958, 180/1.838, 135/2.993} {
    \fill[rlcolor] (\angle:\rr) circle (2pt);
  }
  \draw[basecolor, thick] (-2.2,-4.5) -- (-1.6,-4.5);
  \fill[basecolor] (-1.9,-4.5) circle (2pt);
  \node[right, font=\scriptsize] at (-1.6,-4.5) {Qwen-Image-2.0-Base};
  \draw[rlcolor, thick] (-2.2,-5.0) -- (-1.6,-5.0);
  \fill[rlcolor] (-1.9,-5.0) circle (2pt);
  \node[right, font=\scriptsize] at (-1.6,-5.0) {Qwen-Image-2.0-RL};
\end{tikzpicture}
\caption{Elo rating profile across categories.}
\label{fig:arena_radar}
\end{subfigure}
\hfill
\begin{subfigure}[b]{0.50\textwidth}
\centering
\begin{tikzpicture}
\begin{axis}[
    ybar,
    bar width=4.5pt,
    width=\textwidth,
    height=6cm,
    symbolic x coords={Overall, Product, 3D, Cartoon, Photorealism, Art, Portraits, Text Rendering, Edit},
    xtick=data,
    x tick label style={rotate=45, anchor=east, font=\scriptsize},
    ymin=950, ymax=1400,
    ylabel={Elo Rating},
    ylabel style={font=\scriptsize},
    y tick label style={font=\scriptsize},
    ytick={1000, 1100, 1200, 1300, 1400},
    legend style={at={(0.02,0.98)}, anchor=north west, font=\scriptsize},
    enlarge x limits=0.08,
    grid=major,
    grid style={gray!20},
]
\addplot[fill=basecolor, draw=basecolor!80] coordinates {
    (Overall,1115) (Product,1116) (3D,1093) (Cartoon,1107)
    (Photorealism,1129) (Art,1109) (Portraits,1138) (Text Rendering,1117) (Edit,1256)};
\addplot[fill=rlcolor, draw=rlcolor!80] coordinates {
    (Overall,1193) (Product,1182) (3D,1186) (Cartoon,1186)
    (Photorealism,1220) (Art,1172) (Portraits,1211) (Text Rendering,1195) (Edit,1349)};
\legend{Qwen-Image-2.0-Base,Qwen-Image-2.0-RL}
\end{axis}
\end{tikzpicture}
\caption{Grouped Elo rating comparison.}
\label{fig:arena_bar}
\end{subfigure}
\caption{Human preference evaluation on arena. RL training consistently improves Elo ratings across all eight sub-categories and the overall score.}
\label{fig:arena}
\end{figure}
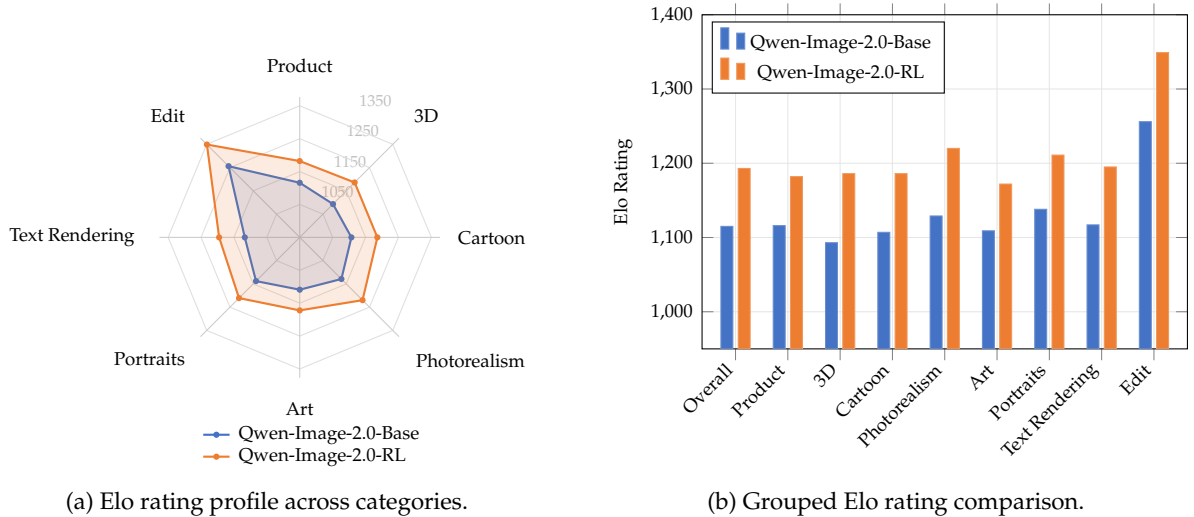

\paragraph{Human preference evaluation.}
Beyond automated benchmarks, we evaluate Qwen-Image-2.0-RL on text-to-image and image edit arena, where users vote between anonymized image pairs from competing models. \cref{fig:arena} presents the Elo ratings of Qwen-Image-2.0 before and after RL training across eight sub-categories. RL training yields substantial Elo rating improvements across all dimensions, with the overall T2I rating rising from 1115 to 1193 (+78). The most pronounced gains appear in 3D Modeling (+93), followed by Photorealism (+91), reflecting improved structural consistency and fine-grained detail rendering. Consistent with the T2I evaluation above, the performance in image edit arena also improves from 1256 to 1349 (+93), further corroborating the effectiveness of our editing-specific RL training.

\section{Related Works}

\paragraph{Reinforcement learning for diffusion models.}
Aligning diffusion models with human preferences via RL has gained significant attention. Flow-GRPO~\citep{liu2025flowgrpo} extends GRPO to flow matching models by treating the multi-step generation as a Markov decision process. GRPO-Guard~\citep{wang2025grpo} identifies implicit over-optimization issues in flow matching RL and proposes regulated clipping to mitigate them. AWM~\citep{xue2025advantage} and DiffusionNFT~\citep{zheng2025diffusionnft} introduces online reinforcement learning with a forward process formulation. Our work builds upon these methods, introduce a hybrid CFG strategy specifically designed for the Qwen-Image-2.0 architecture.

\paragraph{Reward models for diffusion models.}
Reward modeling is central to aligning diffusion models with human preferences. Early CLIP-based approaches~\citep{clip} are limited by fixed architectures and poor cross-task generalization. Subsequent regression-based methods ImageReward~\citep{xu2023imagereward}, PickScore~\citep{kirstain2023pick}, HPSv2~\citep{wu2023human}, and HPSv3~\citep{ma2025hpsv3} train scalar predictors via Bradley--Terry ranking loss. More recently, UnifiedReward~\citep{wang2025unified,wang2026unified} and RewardDance~\citep{wu2025rewarddance} reformulate scoring as token prediction probability, enabling scaling along both model size and context dimensions including task-specific instructions and chain-of-thought (CoT) reasoning. 

\paragraph{On-policy distillation.}
On-policy distillation (OPD) consolidates heterogeneous capabilities by having students learn on self-generated trajectories under teacher supervision. GKD~\citep{agarwal2024policy} established this framework for LLMs, and frontier models have since adopted multi-teacher OPD to avoid the seesaw effect of multi-reward RL. Two concurrent works Flow-OPD~\citep{fang2026flow} and DiffusionOPD~\citep{li2026diffusionopd} extend OPD to flow matching models by showing that the KL divergence between Gaussian transition kernels reduces to a velocity-field MSE loss. Both of these works focus on consolidating single-reward T2I teachers. Our approach differs by specializing teachers by task type (T2I and editing), deriving the objective from a $W_2$ upper bound.

\section{Conclusion}

We presented Qwen-Image-2.0-RL, an image generation system that combines RLHF with OPD to substantially improve visual quality and instruction-following capabilities. Our approach makes three contributions: (1) a VLM-based composite reward system tailored to both T2I and TI2I tasks, with structured evaluation rubrics covering aesthetic quality, prompt adherence, portrait fidelity, instruction-following, and visual consistency; (2) an adapted GRPO training framework featuring a hybrid CFG strategy and asynchronous reward pipeline that enables efficient large-scale RL training of flow matching models; and (3) an OPD mechanism that unifies task-specialized RL-trained teachers into a single deployment model via trajectory-level velocity matching, eliminating reward model dependency while preserving the quality gains of each specialized teacher. Qualitative and quantitative evaluations confirm that OPD not only matches but surpasses a mixed RL baseline that jointly optimizes all task rewards, validating the decomposed training strategy. The resulting system significantly improves performance across human preference benchmarks for both image generation and editing tasks.

\clearpage
\appendix

\section{Derivation of the On-Policy Distillation Objective}
\label{app:opd_derivation}

In this section, we provide a formal derivation of the OPD loss (\cref{eq:opd_loss}) from the perspective of distributional distance minimization.

\paragraph{Goal.} Let $\mathbf{v}_{\theta^\star}$ denote a task-specialized RL-trained teacher and $\mathbf{v}_\theta$ a student model (initialized from the pre-trained base model). Both share the same initial noise distribution $p_1 = \mathcal{N}(\mathbf{0}, \mathbf{I})$ and generate images by solving the reverse ODE (\cref{eq:sampling-ode}) from $t=1$ to $t=0$. The goal of OPD is to find $\theta$ such that the student's output distribution $p_0^\theta$ is as close as possible to the teacher's $p_0^{\theta^\star}$.

\paragraph{Analogy to LLM distillation.} In large language model (LLM) distillation, the standard approach minimizes the forward Kullback--Leibler (KL) divergence between the student's and teacher's output distributions:
\begin{align}
    \min_\theta \; \operatorname{KL}\!\left(p_\theta(\cdot | c) \,\|\, p_{\theta^*}(\cdot | c)\right),
\end{align}
where $p_\theta$ and $p_{\theta^*}$ are the student's and teacher's next-token distributions conditioned on context $c$. The KL divergence admits a tractable decomposition over the autoregressive factorization, making it a natural choice for sequential discrete models. For continuous-space diffusion models, however, the output distributions are defined implicitly through an ODE solve, and the KL divergence between path measures is generally intractable.

\paragraph{Wasserstein-2 distance.} We instead consider the 2-Wasserstein distance as the distributional metric:
\begin{align}
    W_2(p_0^\theta, p_0^{\theta^*}) = \left(\inf_{\gamma \in \Gamma(p_0^\theta, p_0^{\theta^*})} \int \|\mathbf{x} - \mathbf{y}\|^2 \,\mathrm{d}\gamma(\mathbf{x}, \mathbf{y})\right)^{1/2},
\end{align}
where $\Gamma(p_0^\theta, p_0^{\theta^*})$ denotes the set of all couplings between the two distributions. However, directly computing $W_2$ distance requires solving an optimal transport problem, which is intractable in high dimensions. We therefore seek to minimize an efficiently computable upper bound.

\citet{benton2023error} establish a rigorous connection between the velocity field approximation error and the distributional distance of the corresponding flows. Their result relies on the following assumptions:

\begin{enumerate}[leftmargin=*, itemsep=2pt, label=\textbf{(A\arabic*)}]
    \item \emph{Existence and uniqueness of smooth flows.} For each initial point $\mathbf{x}_1 \in \mathbb{R}^d$, both the student flow (under $\mathbf{v}_\theta$) and the teacher flow (under $\mathbf{v}_{\theta^*}$) admit unique, continuously differentiable solutions.
    \item \emph{Lipschitz regularity of teacher velocity field.} For each $t \in (0, 1)$, there exists a constant $L_t$ such that $\mathbf{v}_{\theta^\star}(\cdot, t)$ is $L_t$-Lipschitz in $\mathbf{x}$.
\end{enumerate}

Under these assumptions, Theorem 1 of \citet{benton2023error} yields the following bound on the $W_2$ distance between the student's and teacher's generated distributions:
\begin{align}
    W_2(p_0^\theta, \, p_0^{\theta^\star}) \;\leq\; \left(\int_0^1 \mathbb{E}_{\mathbf{x}_t \sim p_t^{\theta}} \left\| \mathbf{v}_\theta(\mathbf{x}_t, t) - \mathbf{v}_{\theta^*}(\mathbf{x}_t, t) \right\|^2 \mathrm{d}t\right)^{\frac{1}{2}} \cdot \exp\!\left(\int_0^1 L_t \,\mathrm{d}t\right).
    \label{eq:w2_bound}
\end{align}

\paragraph{Continuous-time optimization objective.} The exponential factor $\exp\!\left(\int_0^1 L_t \,\mathrm{d}t\right)$ depends on the Lipschitz regularity of the teacher's velocity field, which is controlled by the network architecture and does not depend on the training objective. Therefore, minimizing the $W_2$ upper bound in \cref{eq:w2_bound} with respect to $\theta$ reduces to minimize the following trajectory-level velocity matching objective:
\begin{align}
   \mathcal{L}_{\text{OPD}}(\theta) = \int_0^1 \mathbb{E}_{\mathbf{x}_t \sim p_t^{\theta}} \left\| \mathbf{v}_\theta(\mathbf{x}_t, t) - \mathbf{v}_{\theta^*}(\mathbf{x}_t, t) \right\|^2 \mathrm{d}t.
    \label{eq:opd_continuous}
\end{align}
In practice, the student's ODE is solved with $N$ discrete steps using a timestep schedule $\{t_0 = 1, t_1, \ldots, t_N \approx 0\}$. The student's own trajectory $\{\mathbf{x}_{t_0}, \mathbf{x}_{t_1}, \ldots, \mathbf{x}_{t_N}\}$ provides samples from $p_{t_n}^\theta$ at each timestep. Approximating the continuous integral in \cref{eq:opd_continuous} by a sum over the discrete trajectory points recovers the practical OPD loss:
\begin{align}
    \mathcal{L}_{\text{OPD}}(\theta) = \mathbb{E}_{c,\mathbf{x}_{[1:N]}\sim \pi_\theta(\cdot|c)}\left[\sum_{n =1}^N \left\Vert \mathbf{v}_\theta(\mathbf{x}_{t_n}, t_n) - \mathbf{v}_{\theta^*}(\mathbf{x}_{t_n}, t_n) \right\Vert^2\right].
\end{align}
This is precisely the training objective in \cref{eq:opd_loss}.

\bibliography{colm2024_conference}
\bibliographystyle{colm2024_conference}

\end{document}